\documentclass{ceurart} 

\usepackage{latexsym}
\usepackage{amssymb}
\usepackage{amsmath}
\usepackage{amsthm}
\usepackage{booktabs}
\usepackage{enumitem}
\usepackage{graphicx}
\usepackage{tabularx}
\usepackage{color}
\usepackage{multirow}
\usepackage{array}  
\usepackage{ragged2e}
\usepackage{hyperref}
\usepackage{caption}
\usepackage{float}
\usepackage{placeins}
\usepackage{adjustbox}
\usepackage{longtable}

\newcommand{\BibTeX}{B\kern-.05em{\sc i\kern-.025em b}\kern-.08em\TeX}
\newcolumntype{Y}{>{\raggedright\arraybackslash}X}

\begin{document}

\copyrightyear{2025}
\copyrightclause{Copyright for this paper by its authors. Use permitted under Creative Commons License Attribution 4.0 International (CC BY 4.0).}

\conference{Accepted for presentation at AEQUITAS 2025: Workshop on Fairness and Bias in AI | co-located with ECAI, \\ October 26th, 2025, Bologna, Italy}

\title{Mechanistic Interpretability with SAEs: Probing Religion, Violence, and Geography in Large Language Models}

\author[1]{Katharina Simbeck}[%
orcid=0000-0001-6792-461X,
email=katharina.simbeck@htw-berlin.de,
url=https://iug.htw-berlin.de,
]
\cormark[1]
\fnmark[1]

\author[1]{Mariam Mahran}[%
orcid=0009-0003-0568-0172,
email=mariam.mahran@htw-berlin.de,
url=https://mariamkhmahran.github.io/connect/,
]
\cormark[1]
\fnmark[1]

\address[1]{HTW Berlin University of Applied Sciences, Treskowallee 8, 10318 Berlin, Germany}
\cortext[1]{Corresponding author}
\fntext[1]{These authors contributed equally.}

%%%%%%%%%%%%%%%%%%%%%%%%%%%%%%%%%%%%%%%%%%%%%%%%%%%%%%%%%%%%%%%%%%%%%%%%

\begin{abstract}
Despite growing research on bias in large language models (LLMs), most work has focused on gender and race, with little attention to religious identity. This paper explores how religion is internally represented in LLMs and how it intersects with concepts of violence and geography. Using mechanistic interpretability and Sparse Autoencoders (SAEs) via the Neuronpedia API, we analyze latent feature activations across five models. We measure overlap between religion- and violence-related prompts and probe semantic patterns in activation contexts. While all five religions show comparable internal cohesion, Islam is more frequently linked to features associated with violent language. In contrast, geographic associations largely reflect real-world religious demographics, revealing how models embed both factual distributions and cultural stereotypes. These findings highlight the value of structural analysis in auditing not just outputs but also internal representations that shape model behavior.
\end{abstract}


\maketitle

%%%%%%%%%%%%%%%%%%%%%%%%%%%%%%%%%%%%%%%%%%%%%%%%%%%%%%%%%%%%%%%%%%%%%%%%

\section{Introduction}
The rapid rise of large language models (LLMs) has transformed natural language processing but also raised serious concerns about embedded biases \cite{10.1162/coli_a_00524, plaza-del-arco-etal-2024-divine}. Trained on extensive datasets often sourced from the internet, LLMs often mirror (and even amplify) existing stereotypes in the data \cite{oketunji2023large}. If left unchecked, these biases can lead to discriminatory outcomes, especially in sensitive domains such as education, recruitment, and information dissemination \cite{Simbeck2024}. Mechanistic interpretability offers a powerful lens for uncovering hidden conceptual structures within LLMs \cite{Conmy2023, galichin2025icoveredbaseshere, lieberum2024gemma, shu2025surveysparseautoencodersinterpreting}. Sparse Autoencoders (SAEs), in particular, have been introduced as a method for mechanistically interpreting LLMs \cite{pmlr-v202-chughtai23a}. By enforcing sparsity in the feature space, SAEs enable the identification of individual activations linked to meaningful concepts. This approach not only reveals internal patterns of bias, but also opens the door for targeted mitigation through adjustments to latent feature weights.

Mainstream LLM bias research has prioritized aspects like gender and race, with religion receiving less attention \cite{10.1162/coli_a_00524, li2023survey}. Prior work shows that religion is a sensitive axis for stereotyping \cite{plaza-del-arco-etal-2024-divine, abid, abrar2025religiousbiaslandscapelanguage}, yet few studies have explored how such biases are internalized in LLMs’ latent spaces.

\vspace{1ex}
\noindent \textbf{To guide our analysis, we consider the following questions:} \vspace{-1.5ex}
\begin{itemize}
    \item \textbf{RQ1:} How consistently do LLMs encode each religion as a distinct and coherent latent concept?
    \vspace{-1.5ex}
    \item \textbf{RQ2:} To what extent are religious identities internally associated with violence in LLMs?
    \vspace{-1.5ex}
    \item \textbf{RQ3:} To what degree do LLMs encode geographic patterns of religion, and how closely do these patterns reflect real-world distributions?
    \vspace{-1.5ex}
    \item \textbf{RQ4:} How do these associations vary across model architectures and training datasets?
\end{itemize}
\vspace{-1ex}

Using the Neuronpedia API \cite{neuronpedia23Lin}, we analyzed latent feature activations across five models using two forms of analysis:  (1) intra-group overlap: how consistently religion-related prompts activate shared features within each model (\textbf{RQ1}); and (2) inter-group overlap: the extent to which religion-related features overlap with violent-related features (\textbf{RQ2}).
We also perform semantic probing on the most-activated features to identify patterns tied to violence (\textbf{RQ2}) and geographic associations (\textbf{RQ3}). Differences across models and datasets are used to address \textbf{RQ4}.

%%%%%%%%%%%%%%%%%%%%%%%%%%%%%%%%%%%%%%%%%%%%%%%%%%%%%%%%%%%%%%%%%%%%%%%%

\section{Related work: Explainability and Interpretability of LLMs}

\subsection{Early vs. New Interpretability Techniques}

Biases in LLMs can be broadly categorized into \textbf{intrinsic biases}, embedded in internal representations, and \textbf{extrinsic biases}, which manifest in generated outputs \cite{li2023survey}. Due to the scale of modern models, intrinsic biases are more difficult to study. Most existing research focuses on output-level evaluations, using word association tests or probability-based templates \cite{caliskan, may2019measuringsocialbiasessentence, Kaneko_Bollegala_2022, kurita2019measuringbiascontextualizedword, nadeem2020stereosetmeasuringstereotypicalbias}.

Moving beyond output-based analysis, researchers have explored how models represent information internally. Two common early techniques include attention head visualization and neuron-level interpretation. The former highlights token relationships across layers, but often fails to reflect the true basis of model decisions \cite{vig2019multiscalevisualizationattentiontransformer, wiegreffe2019attentionexplanation}. The latter examines whether individual neurons encode meaningful concepts, although interpretations often lack consistency across datasets \cite{Dalvi_Durrani_Sajjad_Belinkov_Bau_Glass_2019, bolukbasi2021interpretabilityillusionbert}.

Given the limitations of earlier methods, more recent work has shifted toward \textbf{mechanistic interpretability} as a way to gain deeper insight into model internals. Rather than treating models as black boxes, this approach aims to make sense of the finite, yet massive, number of model parameters \cite{shu2025surveysparseautoencodersinterpreting, olah}. This technique offers a more structured way to investigate how concepts such as identity, belief, or bias are internally encoded. One major challenge for mechanistic interpretability is neuron polysemanticity, where individual neurons are activated by multiple, often unrelated, concepts across different contexts \cite{cunningham2023sparseautoencodershighlyinterpretable}. \textbf{Sparse autoencoders} have been proposed to address this issue by learning new, disentangled features from model activations \cite{galichin2025icoveredbaseshere, shu2025surveysparseautoencodersinterpreting, cunningham2023sparseautoencodershighlyinterpretable}. The key idea is that, by enforcing sparsity, only a few features are active for a given input, making it easier to isolate individual concepts to interpret them.

SAEs are shallow neural networks with a single hidden layer, trained to reconstruct LLM activations under a sparsity constraint. The encoder identifies active features for a given input, and the decoder maps the features back to the original state \cite{he2024llama}. The resulting sparse features (\textbf{latent features}) are hypothesized to correspond to real semantically meaningful concepts embedded within the model \cite{lieberum2024gemma}. SAEs are trained on LLM activations recorded at specific points in the transformer block, using tokenized input text. These activations form the basis for learning a sparse feature dictionary \cite{rajamanoharan2024jumpingaheadimprovingreconstruction}.

\subsection{Religious Bias in LLMs}

Bias in LLMs has been widely studied in dimensions such as gender, race, and political ideology, with many tools developed to assess and mitigate these issues \cite{10.1162/coli_a_00524, li2023survey}. However, religious bias remains underexplored despite its potential to reinforce stereotypes and marginalize vulnerable communities. Some studies have shown that models associate “Muslim” with “terrorist” or “Jewish” with “money” \cite{abid}, and that Islam is more frequently linked to violence than Christianity in both language and text-to-image models \cite{abrar2025religiousbiaslandscapelanguage}. Others found that western religions are represented with more nuance, while eastern ones are reduced to oversimplified representations \cite{plaza-del-arco-etal-2024-divine}. These findings show that religion is a sensitive axis of model bias but one that has received less attention in mainstream research. Addressing this gap is essential for developing AI systems that respect and fairly represent diverse religious identities.

\begin{table*}[t]
\caption{Summary of SAE configurations used in this study, including model architecture, number of features, training corpus, and relevant access links~\cite{lieberum2024gemma, he2024llama, bloom2024gpt2residualsaes}. All SAEs are accessed through Neuronpedia \cite{neuronpedia23Lin}.}
\label{tab:sae_summary}
\scalebox{0.98}{
\resizebox{\textwidth}{!}{%
\begin{tabular}{p{3.4cm} p{2.4cm}p{1.4cm} p{3.3cm} p{7cm}}
\toprule
\textbf{Short Name} & \textbf{LLM} & \textbf{\#Features} & \textbf{Training Corpus} & \textbf{Reference} \\
\midrule
res-jb & GPT-2 Small& 24,576 & OpenWebText & \url{https://www.neuronpedia.org/gpt2-small/res-jb} \\
att-kk & GPT-2 Small& 24,576 & OpenWebText & \url{https://www.neuronpedia.org/gpt2-small/att-kk} 
\textit{(under peer review)} \\
gemmascope-att-16k & Gemma-2-2B & 16,384 & Pile-Uncopyrighted & \url{https://www.neuronpedia.org/gemma-2-2b/gemmascope-att-16k} \\
gemmascope-res-16k & Gemma-2-2B & 16,384 & Pile-Uncopyrighted & \url{https://www.neuronpedia.org/gemma-2-2b/gemmascope-res-16k} \\
gemmascope-res-65k & Gemma-2-2B & 65,536 & Pile-Uncopyrighted & \url{https://www.neuronpedia.org/gemma-2-2b/gemmascope-res-65k} \\
gemmascope-res-16k & Gemma-2-9B & 16,384 & Pile-Uncopyrighted & \url{https://www.neuronpedia.org/gemma-2-2b/gemmascope-res-16k} \\
gemmascope-res-16k & Gemma-2-9B-IT & 16,384 & Pile-Uncopyrighted & \url{https://www.neuronpedia.org/gemma-2-9b-it/gemmascope-res-16k} \\
gemmascope-res-131k & Gemma-2-9B-IT & 131,072 & Pile-Uncopyrighted & \url{https://www.neuronpedia.org/gemma-2-9b-it/gemmascope-res-131k} \\
llamascope-res-32k & Llama3.1-8B & 32,768 & SlimPajama & \url{https://huggingface.co/fnlp/Llama-Scope} \\
\bottomrule
\end{tabular}}}
\vspace{-3ex}
\end{table*}

%%%%%%%%%%%%%%%%%%%%%%%%%%%%%%%%%%%%%%%%%%%%%%%%%%%%%%%%%%%%%%%%%%%%%%%%

\section{Data Collection}

This analysis relies on the Neuronpedia API\footnote{\url{https://www.neuronpedia.org}} to access sparse, interpretable feature representations extracted from a variety of LLMs using SAEs. The platform provides pre-trained SAEs across multiple model families, allowing for detailed analysis of internal model representations. While we do not revalidate the underlying SAEs here, they are drawn from prior peer-reviewed work \cite{lieberum2024gemma, neuronpedia23Lin, he2024llama, bloom2024gpt2residualsaes}. We treat them as a reasonable foundation for exploring internal associations.

The API returns the top activating latent features for each query, along with metadata such as feature ID, activation layer, and highly activating example texts. Table~\ref{tab:sae_summary} summarizes the SAEs used in this study, which include variations of GPT2-small, Gemma-2, and Llama3.1-8B. These were trained at both attention (\texttt{-att}) and residual stream (\texttt{-res}) positions \cite{lieberum2024gemma, he2024llama, bloom2024gpt2residualsaes}.

To collect data, we constructed a set of minimal, controlled natural language prompts targeting religious and violence-related concepts.\footnote{Code and data are available at \url{https://github.com/iug-htw/SAE_fairness}.} Each was submitted to the Neuronpedia API, and the top 20 activating feature IDs were stored per model, along with the most activating texts for later semantic analysis. We focus on five major world religions: Christianity, Islam, Judaism, Hinduism, and Buddhism. For each, we curated representative keywords (e.g., sacred texts, places of worship) and embedded them in simple declarative sentences with consistent structure like “This is the Quran” or “This is a church.”  Table~\ref{tab:religion_terms} (Appendix~\ref{appendix:religion_terms_feat_overlap}) lists all religious terms used.
We also created a smaller set of prompts related to violence and criminality (e.g., “terrorist,” “extremist”). These were used to assess overlap between religion- and violence-activated features.

%%%%%%%%%%%%%%%%%%%%%%%%%%%%%%%%%%%%%%%%%%%%%%%%%%%%%%%%%%%%%%%%%%%%%%%%

\section{Data Analysis}

\subsection{Latent Feature Overlap Analysis}

We conducted a two-part analysis that focused on latent feature overlaps. We explore whether certain religions activate coherent internal structures within a model (intra-group overlap) and whether these structures intersect with violence-related concepts (inter-group overlap). The main goal is to assess representation consistency, as well as potential internalized associations with harmful stereotypes.

\subsubsection{Intra-group Overlap}

To test how consistently LLMs encode each religion as a coherent concept (\textbf{RQ1}), we measured how often prompts for the same religion activated the same latent features. Higher overlap means that the model tends to treat the religion as one clear idea, instead of spreading the activations across a wide feature space. We initially explored cosine similarity between binary feature vectors, but the results were too noisy due to extreme sparsity, with some prompts activating less than 20 out of over 100,000 features. Discrete counts provided more interpretable and stable results and were used instead.

As shown in Table~\ref{tab:feature_overlap_summary}, all five religions show similar intra-group overlap across models. For example, in GPT2-small, Buddhism and Hinduism each share about 60 features across their respective prompts. In Gemma-2-9b-IT, overlap ranges from 36 (Islam) to 43 (Buddhism), suggesting comparable internal consistency across religious categories.

We also computed the total number of unique features activated by all religious prompts combined (regardless of group), to assess the overall compactness of religious representations. The results ranged between 18 to 145 across models. Gemma-2-9b had the most compact representation, indicating it encodes religion using a small, overlapping set of features. This potentially reflects efficient abstraction, but also less differentiation between religions.

While these results don’t reveal strong religious bias, they show that all five religions are represented with similar structural cohesion \textbf{RQ1}. This serves as a useful baseline to ensure that differences observed in later bias analyses are not simply side effects of representation inconsistency.

\begin{table*}[t]
\caption{Latent feature overlap by query, category, and model. Upper section shows raw intra-group overlaps (within religions); lower section shows raw inter-group overlaps with violence-related prompts, along with the Violence Association Index (VAI). VAI expresses each value as a percentage of the model's average across religions (100 = model average).}

\label{tab:feature_overlap_summary}
\centering
\footnotesize
\setlength{\tabcolsep}{8.5pt} 
\scalebox{0.95}{
\begin{tabular}{lrrrrrrrrrrr}
\toprule
\textbf{Category} &
\multicolumn{2}{c}{Gemma-2-2b} &
\multicolumn{2}{c}{Gemma-2-9b} &
\multicolumn{2}{c}{Gemma-2-9b-IT} &
\multicolumn{2}{c}{GPT2-small} &
\multicolumn{2}{c}{Llama3.1-8B} \\
\midrule
\multicolumn{11}{l}{\textbf{Intra-group Overlap}} \\
\midrule
Christianity            & 40 &      & 17 &      & 40 &      & 51 &      & 13 &      \\
Islam                   & 39 &      & 16 &      & 36 &      & 46 &      & 11 &      \\
Judaism                 & 37 &      & 17 &      & 37 &      & 45 &      & 11 &      \\
Buddhism                & 43 &      & 17 &      & 43 &      & 60 &      & 16 &      \\
Hinduism                & 37 &      & 16 &      & 41 &      & 61 &      & 19 &      \\
All Religions Combined  & 51 &      & 18 &      &117 &      &145 &      & 43 &      \\
\midrule
\multicolumn{11}{l}{\textbf{Inter-group Overlap with Violence-related Concepts}} \\
\midrule
& Raw & Index & Raw & Index & Raw & Index & Raw & Index & Raw & Index \\
\midrule
Christianity (terrorism)    &  37 &  96 & 17 & 94 & 24 & 94 & 35 & 94 & 11 & 100 \\
Islam (terrorism)           &  \textbf{45} & \textbf{117} & \textbf{22} & \textbf{122} & \textbf{28} & \textbf{110} & \textbf{42} & \textbf{113} & \textbf{12} & \textbf{109} \\
Judaism (terrorism)         &  37 &  96 & 17 & 94 & 27 & 106 & 35 & 94 & 11 & 100 \\
Buddhism (terrorism)        &  37 &  96 & 17 & 94 & 24 & 94 & 36 & 96 & 11 & 100 \\
Hinduism (terrorism)        &  36 &  94 & 17 & 94 & 24 & 94 & 38 & 102 & 10 &  90 \\
\bottomrule
\end{tabular}}
\vspace{-2.5ex}
\end{table*}

\subsubsection{Inter-group Overlap}

The second part of our analysis examined whether internal representations of religion intersect with harmful concepts like “terrorism” (\textbf{RQ2}). We compiled all unique latent features activated per religion group, then compared them to the set of features activated by a group of crime-related prompts. The overlap count reflects how many latent features are shared between the two groups. To enable meaningful comparisons between models, we calculated the \textit{Violence Association Index (VAI)}. This index normalizes the raw overlap by dividing it by the mean overlap across all five religions within the same model, then multiplying by 100. A VAI above 100 indicates stronger-than-average association with violence-related features within that model, while below 100 indicates weaker association.

As shown in Table~\ref{tab:feature_overlap_summary}, Islam consistently registers the highest VAI across all five models. In Gemma-2-2b, for example, Islam scores 117, while other religions range from 94 to 96. A similar pattern appears in GPT2-small (Islam: 113; others range from 94 to 102). Although the raw overlap values for some religions are close, the VAI highlights a consistent relative skew across models.

\subsection{Semantic Activation Analysis}

We examined top activation texts linked to religion-related features to explore whether certain themes appear more often in the internal representations of specific religions. We focus on two semantic categories: crime and geographical representations. Crime-related associations can suggest whether models implicitly link religious identity with violence or extremism. Geographic associations help reveal whether representations align with real-world religious distributions or reflect biased mappings.

The Neuronpedia API returns top activating texts for each feature, offering insight into their semantic content. For each religion and model, we gathered texts related to the top 20 features per query. We then applied a keyword-based search to these texts using predefined keyword lists for crime and geographic terms (Appendix \ref{appendix:religion_terms_sem_analysis}).

\subsubsection{Crime Analysis}

To further address \textbf{RQ2}, we conducted a semantic analysis of activation texts related to each religion, searching for twelve crime-related keywords (e.g., “terrorism,” “extremist,” “crime,” “shooting,” “violence”) within the top texts returned by the Neuronpedia API.

Keyword matches were normalized as a percentage of all activation texts per religion and model. As shown in Table~\ref{tab:crime_mentions}, Islam consistently had the highest proportion of crime-related terms in most models. For example, Islam scored 3.46\% in Gemma-2-9b-IT, compared to 2.44\% for Christianity and 2.20\% for Judaism. However, in GPT2-small and Llama3.1-8B, Hinduism showed unexpectedly higher rates than Islam. These differences may reflect training data co-occurrence patterns shaped by regional sociopolitical discourse. Such variation across models hints that associations between religion and violence depend on architecture and dataset composition (\textbf{RQ4}).

\begin{table*}[t]
\small
\caption{Crime-related activation text percentages by model and religion. Highest crime associations in bold.}
\label{tab:crime_mentions}
\centering
\setlength{\tabcolsep}{4pt} 
\scalebox{0.96}{
\resizebox{\textwidth}{!}{
\begin{tabular}{p{2.3cm}|p{1cm}p{1.3cm}|p{1cm}p{1.3cm}|p{1cm}p{1.3cm}|p{1cm}p{1.3cm}|p{1cm}p{1.3cm}}
\toprule
\textbf{Model} &
\multicolumn{2}{c|}{\textbf{Christianity}} &
\multicolumn{2}{c|}{\textbf{Islam}} &
\multicolumn{2}{c|}{\textbf{Judaism}} &
\multicolumn{2}{c|}{\textbf{Buddhism}} &
\multicolumn{2}{c}{\textbf{Hinduism}} \\
\cmidrule(r){2-3} \cmidrule(r){4-5} \cmidrule(r){6-7} \cmidrule(r){8-9} \cmidrule(r){10-11}
& \textbf{\#Texts} & \textbf{\% Crime}
& \textbf{\#Texts} & \textbf{\% Crime}
& \textbf{\#Texts} & \textbf{\% Crime}
& \textbf{\#Texts} & \textbf{\% Crime}
& \textbf{\#Texts} & \textbf{\% Crime} \\
\midrule
gemma-2-2b    & 23695 & 1.87\% & 23180 & \textbf{2.02\%} & 20993 & 1.82\% & 22486 & 1.73\% & 20541 & 1.87\% \\
gemma-2-9b    & 9675  & 2.77\% & 9630  & \textbf{2.98\%} & 8880  & 2.75\% & 9403  & 2.73\% & 8909  & 2.68\% \\
gemma-2-9b-it & 21477 & 2.44\% & 20575 & \textbf{3.46\%} & 19512 & 2.20\% & 20773 & 2.12\% & 18288 & 2.65\% \\
GPT2-small    & 22284 & 6.43\% & 22046 & 6.88\% & 20568 & 6.74\% & 21950 & 6.72\% & 19954 & \textbf{7.13\%} \\
Llama3.1-8B   & 7383  & 2.52\% & 7862  & 3.23\% & 7253  & 2.54\% & 7551  & 3.26\% & 7530  & \textbf{4.44\%} \\
\bottomrule
\end{tabular}}}
\vspace{-2.5ex}
\end{table*}

\subsubsection{Geographic Analysis}

\begin{figure}[t]
    \centering
    \includegraphics[width=0.9\linewidth]{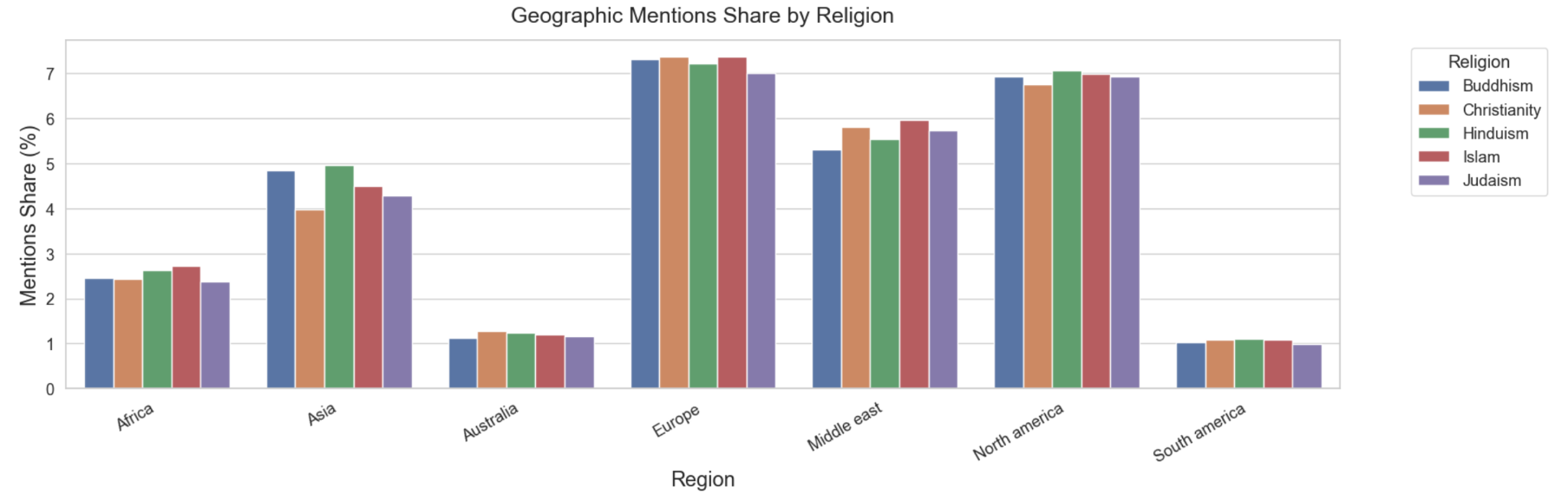}
    \caption{Clustered bar chart showing the number of geographic keyword mentions in activation texts per religion across regions. }
    \label{fig:geo_heatmap}
\vspace{-4ex}
\end{figure}

To address \textbf{RQ3}, we analyzed how LLMs associate religions with geographic regions by scanning activation texts for curated keywords representing seven areas: Africa, Asia, Australia, Europe, the Middle East, North America, and South America. Mentions were counted per religion–region pair. Figure~\ref{fig:geo_heatmap} presents a grouped bar chart of geographic mention shares by religion, illustrating how often different regions appear in the activation texts associated with each religious concept. Europe and North America emerge as the most frequently mentioned regions, with relatively balanced associations among the five religions. Asia and the Middle East also show strong representation, though with more variation. Hinduism and Buddhism dominated the Asian context, while Islam is most prominent in the Middle East. In contrast, Africa and South America exhibit lower overall mention rates and greater disparity between religions, indicating weaker and less consistent associations. Australia had the sparsest coverage, appearing minimally within all religious groups.

Christianity and Islam appear across nearly all regions, reflecting a broad conceptual reach. Hinduism and Buddhism are more localized but still surface meaningfully outside Asia, particularly in the Americas. Judaism, by comparison, has a markedly narrower geographic distribution.

%%%%%%%%%%%%%%%%%%%%%%%%%%%%%%%%%%%%%%%%%%%%%%%%%%%%%%%%%%%%%%%%%%%%%%%%

\section{Discussion}

Our analysis shows that religion in LLMs is not only internally coherent, but also systematically entangled with broader cultural narratives. The implications are twofold: (1) LLMs reliably abstract religion into stable latent categories, and (2) those categories often co-activate with features tied to violence or geography, embedding cultural frames into the models’ conceptual space.

All five religions activated compact feature sets (\textbf{RQ1}), showing that the models treat religion as stable rather than diffuse. But stability is not neutrality. Islam’s stronger overlap with violence-related features (\textbf{RQ2}) shows how coherent categories can still encode stereotypes. This highlights a core risk. Stable representations make linked stereotypes more deeply embedded and harder to remove.

Interestingly, some models, such as GPT2-small and Llama3.1-8B, deviated from this trend, where Hinduism showed the strongest crime associations. This variation reflects training data influences, particularly in regions where religion is intertwined with political conflict. It shows that LLMs do not merely learn language but also reproduce the narratives and frames prevalent in their training corpora.

The geographic analysis (\textbf{RQ3}) revealed both expected mappings (e.g., Hinduism–Asia, Christianity–Europe) and some distortions, with Europe and North America strongly represented across all religions, while Australia and South America were largely absent. This points to a Western-focused lens, where media visibility rather than demographics shapes associations. Judaism’s narrow footprint and Islam’s global spread further reflect how internal representations mirror cultural salience more than statistical reality, with potential downstream impacts on tasks like summarization or retrieval.

The differences across models (\textbf{RQ4}) remind us that interpretability findings cannot be generalized across architectures. Smaller models like GPT2-small revealed noisier and more exaggerated associations (e.g., Hinduism-crime), while larger models like Gemma-2-9b encoded more compact and abstract representations. This suggests that bias is shaped not just by data but by model scale and structure. Model audits therefore require granularity. What looks like a universal association may in fact be contingent on model class.

The key contribution of this work is methodological. By moving beyond outputs, we show how SAEs uncover the “conceptual geography” inside LLMs. This matters because downstream harms do not always stem from what models say, but from how they internally prioritize and structure information. A recommender system, for example, could inherit latent religion–violence associations without ever producing an explicitly biased sentence. More broadly, our findings illustrate how interpretability tools can map the transition point where generalization slips into stereotype.

%%%%%%%%%%%%%%%%%%%%%%%%%%%%%%%%%%%%%%%%%%%%%%%%%%%%%%%%%%%%%%%%%%%%%%%%

\section{Conclusion}
This study investigated how LLMs internally represent religious concepts, focusing on structural and semantic associations with violence and geography. Using Sparse Autoencoders, we analyzed latent feature overlap and activation contexts across models and religions. Results showed consistent internal cohesion but notable asymmetries (especially Islam’s stronger links to violence-related features) and revealed how regional narratives are embedded in model representations.

Future work could extend this analysis to more religions, multilingual models, or other identity-linked concepts. As interpretability tools advance, they will be essential for understanding not just model outputs, but the internal structures that shape them.

%%%%%%%%%%%%%%%%%%%%%%%%%%%%%%%%%%%%%%%%%%%%%%%%%%%%%%%%%%%%%%%%%%%%%%%%

\begin{acknowledgments}
This paper portrays the work carried out in the context of the KIWI project (16DHBKI071) that is generously funded by the Federal Ministry of Research, Technology and Space (BMFTR).

We also gratefully acknowledge the Neuronpedia API, which provided access to SAE activations and feature explanations. Their open infrastructure was essential for the experiments conducted in this study.

\end{acknowledgments}

\section*{Declaration on Generative AI}
During the preparation of this work, the authors used OpenAI’s GPT-4.1 to check grammar and spelling and to enhance the writing style. After using these tools/services, the authors reviewed and edited the content as needed and take full responsibility for the publication’s content.
  
%%%%%%%%%%%%%%%%%%%%%%%%%%%%%%%%%%%%%%%%%%%%%%%%%%%%%%%%%%%%%%%%%%%%%%%%

\bibliography{mybibfile}

%%%%%%%%%%%%%%%%%%%%%%%%%%%%%%%%%%%%%%%%%%%%%%%%%%%%%%%%%%%%%%%%%%%%%%%%

\newpage

\appendix

\section{Appendices: List of Representative Keywords Used in The Latent Feature Overlap Analysis}
\label{appendix:religion_terms_feat_overlap}

To generate activations for the latent feature overlap analysis, each keyword listed in this appendix was wrapped in a simple declarative sentence using a fixed template structure (e.g., “This is a church,” “This is a mosque,” “This is a temple”). This approach was adopted after initial experiments using standalone keywords yielded less stable and interpretable features. By embedding each keyword in an identical sentence frame, we aimed to maintain consistency across prompts while minimizing concept leakage. The use of a neutral and uniform structure helps ensure that any background noise introduced by the sentence is equally distributed across all categories and can be averaged out during analysis.

\begin{table}[h]
  \caption{Religion-Related Query Terms and Potential Biases}
  \label{tab:religion_terms}
  \begin{tabularx}{\textwidth}{lY}
    \toprule
    \textbf{Category} & \textbf{Terms} \\
    \midrule
    Christianity & baptism, bible, christian, church, gospel, pope, sacrament, christianity, jesus, priest, pastor, crucifix, communion \\
    Islam & burka, hijab, mosque, muslim, quran, allah, halal, islam, mecca, imam, ramadan, eid, hajj \\
    Judaism & jew, kippah, synagogue, talmud, torah, judaism, kosher, shabbat, rabbi, menorah, mitzvah, hanukkah \\
    Buddhism & buddhist, buddhism, monastery, tripitaka, pagoda, vihara, vesak, monk, buddha, Sangha, mandala, dharma, stupa \\
    Hinduism & hinduism, hindu, mandir, bhagavad gita, varanasi, diwali, holi, puja, Yajna, murti, moksha, brahma, vedas \\ 
    \midrule
    Potential Bias & terrorist, militant, radical, extremist, attack, bombing, bomb, gun, weapon, terror attack, massacre, shooting \\
    \bottomrule
  \end{tabularx}
\end{table}

%%%%%%%%%%%%%%%%%%%%%%%%%%%%%%%%%%%%%%%%%%%%%%%%%%%%%%%%%%%%%%%%%%%%%%%%

\newpage

\section{Appendices: Predefined Keyword Lists for Semantic Analysis of Crime- and Geographic-related terms}
\label{appendix:religion_terms_sem_analysis}

\begin{longtable}{p{0.18\textwidth} p{0.75\textwidth}}
    \caption{Keyword Lists Used for Indexing Semantic Categories} \\
    \label{tab:religion_sem}
    \textbf{Category} & \textbf{Keywords} \\
    \hline
    \endfirsthead

    \multicolumn{2}{c}%
    {{\bfseries \tablename\ \thetable{} -- continued from previous page}} \\
    \textbf{Category} & \textbf{Keywords} \\
    \hline
    \endhead
    
    \hline \multicolumn{2}{r}{{Continued on next page}} \\
    \endfoot

    \endlastfoot

    Crime & terrorism, terrorist, crime, criminal, violence, extremist, extremism, attack, radical, assault, shooting, bomb \\ 
    \midrule
    \midrule
    Europe & europe, european, eurozone, schengen, western europe, eastern europe, northern europe, southern europe, scandinavia, balkans, benelux, iberian peninsula, baltic states, central europe, united kingdom, britain, england, scotland, wales, northern ireland, france, germany, italy, spain, portugal, netherlands, belgium, switzerland, austria, norway, sweden, denmark, finland, iceland, poland, czech republic, slovakia, hungary, romania, bulgaria, greece, croatia, serbia, bosnia, slovenia, montenegro, albania, macedonia, ukraine, belarus, moldova, russia, georgia, armenia, azerbaijan, ireland, luxembourg, liechtenstein, andorra, monaco, san marino, vatican, london, paris, berlin, rome, madrid, vienna, amsterdam, brussels, oslo, stockholm, copenhagen, helsinki, lisbon, warsaw, prague, budapest, bucharest, sofia, athens, zagreb, belgrade, sarajevo, ljubljana, tirana, skopje, kiev, minsk, chisinau, moscow, tbilisi, yerevan, baku, dublin, luxembourg city, vaduz, andorra la vella, eiffel tower, big ben, colosseum, berlin wall, vatican city, acropolis, buckingham palace, louvre, reichstag, sagrada familia \\ 
    \midrule
    Asia & asia, asian, east asia, south asia, southeast asia, central asia, west asia, indian subcontinent, asian continent, china, india, japan, south korea, north korea, taiwan, mongolia, pakistan, bangladesh, nepal, bhutan, sri lanka, maldives, indonesia, philippines, vietnam, thailand, myanmar, burma, malaysia, singapore, cambodia, laos, brunei, east timor, kazakhstan, uzbekistan, turkmenistan, kyrgyzstan, tajikistan, iran, afghanistan, georgia, armenia, azerbaijan, beijing, shanghai, hong kong, tokyo, osaka, seoul, busan, delhi, mumbai, kolkata, chennai, karachi, lahore, islamabad, dhaka, kathmandu, thimphu, colombo, bangkok, hanoi, ho chi minh, jakarta, manila, kuala lumpur, singapore, phnom penh, vientiane, yangon, tehran, bishkek, dushanbe, ashgabat, almaty, yerevan, tbilisi, baku, great wall, taj mahal, angkor wat, borobudur, meiji shrine, forbidden city, mount fuji, petronas towers, ganges, yangtze, mekong \\ 
    \midrule
    Middle East & middle east, middle eastern, arab, arabic, gulf, levant, arab world, iran, iraq, syria, lebanon, jordan, palestine, gaza, west bank, egypt, saudi arabia, saudi, yemen, oman, emirates, qatar, bahrain, kuwait, turkey, cyprus, sudan, south sudan, libya, mauritania, israel, tel aviv, tehran, baghdad, damascus, aleppo, beirut, amman, jerusalem, gaza city, ramallah, riyadh, mecca, medina, jeddah, sanaa, muscat, doha, dubai, abu dhabi, manama, kuwait city, cairo, giza, alexandria, khartoum, tripoli, ankara, istanbul, nicosia, khartoum, al-azhar, al-aqsa, petra, cedars of lebanon, pyramids, temple mount, jerusalem, sinai \\ 
    \midrule
    Africa & africa, african, sub-saharan africa, north africa, west africa, east africa, central africa, southern africa, saharan, sahel, horn of africa, maghreb, morocco, algeria, tunisia, nigeria, ethiopia, egypt, south africa, kenya, algeria, morocco, sudan, angola, ghana, mozambique, madagascar, cameroon, côte d’ivoire, ivory coast, niger, burkina faso, mali, malawi, zambia, zimbabwe, tanzania, rwanda, uganda, chad, senegal, tunisia, libya, botswana, namibia, lesotho, eswatini, gabon, congo, democratic republic of congo, drc, somalia, south sudan, sierra leone, liberia, benin, togo, djibouti, equatorial guinea, central african republic, gambia, guinea, guinea-bissau, mauritania, seychelles, comoros, cape verde, lagos, abuja, nairobi, addis ababa, johannesburg, cape town, durban, tunis, algiers, accra, dakar, kampala, harare, lusaka, windhoek, gaborone, maputo, antananarivo, kinshasa, brazzaville, bamako, ouagadougou, freetown, bujumbura, mogadishu, djibouti city, sahara desert, kilimanjaro, victoria falls, ngorongoro crater, nile river, zambezi \\ 
    \midrule
    North America & north america, north american, american, the us, usa, united states, u.s., u.s.a., the states, canada, canadian, mexico, mexican, washington, new york, new york city, los angeles, chicago, san francisco, boston, miami, dallas, houston, atlanta, seattle, philadelphia, detroit, phoenix, denver, las vegas, orlando, san diego, austin, toronto, vancouver, montreal, ottawa, calgary, edmonton, quebec city, winnipeg, halifax, mexico city, guadalajara, monterrey, tijuana, cancun, puebla, merida, white house, statue of liberty, niagara falls, grand canyon, empire state building, times square, hollywood, pentagon, capitol hill, liberty bell, alamo, mount rushmore, silicon valley \\ 
    \midrule
    South America & south america, south american, latin america, latinoamerica, andes, argentina, brazil, chile, peru, colombia, venezuela, ecuador, bolivia, paraguay, uruguay, guyana, suriname, french guiana, buenos aires, rosario, cordoba, rio de janeiro, são paulo, brasilia, salvador, recife, santiago, valparaíso, lima, cusco, bogotá, medellín, caracas, quito, la paz, sucre, asunción, montevideo, georgetown, paramaribo, cayenne, amazon rainforest, andes mountains, machu picchu, iguazu falls, cristo redentor, atacama desert, pampas, patagonia, galápagos islands \\ 
    \midrule
    Australia & australia, australian, oceania, australasia, new zealand, kiwi, aotearoa, papua new guinea, fiji, samoa, tonga, vanuatu, solomon islands, micronesia, palau, marshall islands, nauru, tuvalu, kiribati, sydney, melbourne, canberra, brisbane, perth, adelaide, hobart, darwin, auckland, wellington, christchurch, hamilton, dunedin, suva, port moresby, nuku'alofa, great barrier reef, uluru, ayers rock, outback, tasmania, kangaroo island, aboriginal, maori, dreamtime, coral sea, southern ocean, tasman sea, anzac, trans-tasman \\
    \bottomrule
\end{longtable}

\end{document}